RESEARCH ARTICLE  Open Access

# EvoGrader: an online formative assessment tool for automatically evaluating written evolutionary explanations

Kayhan Moharreri[1], Minsu Ha[2] and Ross H Nehm[3*]

## Abstract

EvoGrader is a free, online, on-demand formative assessment service designed for use in undergraduate biology classrooms. EvoGrader's web portal is powered by Amazon's Elastic Cloud and run with LightSIDE Lab's open-source machine-learning tools. The EvoGrader web portal allows biology instructors to upload a response file (.csv) containing unlimited numbers of evolutionary explanations written in response to 86 different ACORNS (Assessing COntextual Reasoning about Natural Selection) instrument items. The system automatically analyzes the responses and provides detailed information about the scientific and naive concepts contained within each student's response, as well as overall student (and sample) reasoning model types. Graphs and visual models provided by EvoGrader summarize class-level responses; downloadable files of raw scores (in .csv format) are also provided for more detailed analyses. Although the computational machinery that EvoGrader employs is complex, using the system is easy. Users only need to know how to use spreadsheets to organize student responses, upload files to the web, and use a web browser. A series of experiments using new samples of 2,200 written evolutionary explanations demonstrate that EvoGrader scores are comparable to those of trained human raters, although EvoGrader scoring takes 99% less time and is free. EvoGrader will be of interest to biology instructors teaching large classes who seek to emphasize scientific practices such as generating scientific explanations, and to teach crosscutting ideas such as evolution and natural selection. The software architecture of EvoGrader is described as it may serve as a template for developing machine-learning portals for other core concepts within biology and across other disciplines.

## Background

The landscape of science assessment is being transformed throughout the educational hierarchy in the United States (National Research Council 2012). The movement towards the assessment of complex tasks—such as explaining one's reasoning, constructing an argument, testing hypotheses, or defending an explanation—is becoming the new norm. The 2012 *Advanced Placement* (AP) biology exam, for example, doubled the number of open-ended questions and dramatically reduced the number of multiple-choice questions (Duncan 2013). The new *Framework for K-12 Science Education* (National Research Council 2012) recommended assessing students' scientific competencies by examining their mastery of so-called scientific practices such as explanation, argumentation, and communication of core ideas, none of which can be meaningfully measured using multiple-choice tests (Nehm, Ha and Mayfield 2012b). The U.S. Secretary of Education notes that every year more and more assessments in K-12 education will move to an online environment, opening up additional opportunities for novel forms of knowledge and skill measurement (Duncan 2013).

In introductory biology—one of the most highly enrolled science classes in the U.S.—formative and summative multiple-choice assessments (e.g., clicker questions, midterm exams) remain the norm, greatly limiting the range of skills and competencies that can be assessed (Ha et al. 2011; Nehm and Haertig 2012). Open-ended formative and

* Correspondence: ross.nehm@stonybrook.edu
[3]Center for Science and Mathematics Education, and Department of Ecology and Evolution, Stony Brook University, 092 Life Sciences Building, 11794 Stony Brook, NY, USA
Full list of author information is available at the end of the article summative assessment tools remain scarce in introductory biology because of the time and cost associated with scoring written responses, raising questions about how biology educators teaching large classes will be able to foster the development of the diverse array of skills and practices emphasized in recently developed education reform documents for undergraduate biology education (e.g., *Vision and Change*: American Association for the Advancement of Science 2011).

In response to these challenges, we introduce EvoGrader, one of the first non-commercial machine-learning assessment tools designed for use in introductory biology. We discuss how the assessment system works, provide evidence to support inferences about how scores align with student understanding, and illustrate how it may be used to assess learning of a core concept—natural selection—in the context of the scientific practice of explanation (National Research Council 2012). We also review the architecture of the underlying automated assessment system and discuss how it could be modified to assess student understanding of other core ideas and other scientific practices.

## An online formative assessment tool
### The Web portal

EvoGrader is a free, online, on-demand assessment service. It automatically grades written (typed) evolutionary explanations produced in response to ACORNS assessment items (Assessment of COntextual Reasoning about Natural Selection, Nehm *et al.* 2012a) and "ACORNS-like" assessment items (e.g. questions found in Bishop and Anderson 1990; and many other evolution education studies). All of these instruments contain similar item formats: open-ended questions that ask students to explain how patterns of evolutionary change occurred (from the standpoint of a biologist). Such assessment tasks have been shown to be very useful for understanding student thinking processes (Opfer et al. 2012) and measuring evolutionary understanding (Nehm and Schonfeld 2008, 2010). ACORNS (and ACORNS-like) items are also useful for high school and undergraduate educators because they allow students to practice building scientific explanations, communicating their ideas through writing, and documenting their understanding of the core idea of natural selection (Demastes et al. 1995). Despite their usefulness, these types of assessment tasks take large amounts of time to score. EvoGrader was developed to solve this problem.

EvoGrader is a web portal (see www.evograder.org) that contains an on-demand query box requesting the upload of students' written evolutionary explanations (in .csv format). After upload and "one-click" analysis, EvoGrader generates reports illustrating the presence or absence of accurate scientific ideas (key concepts), non-normative concepts (naïve ideas), and holistic reasoning models (pure scientific, "mixed", and pure non-normative) in each student's response and in the sample of responses as a whole. The system rapidly performs the difficult work of grading written items that for the past 30 years have been used in science education research and practice (for a review, see Nehm and Ha 2011). Responses to 86 different items can be scored using the EvoGrader web portal, which relies on machine-learning methods.

### Supervised machine learning: the core of EvoGrader

EvoGrader uses machine learning methods to extract key concept scores, naïve idea scores, and holistic reasoning model scores from text responses. An integral part of supervised machine learning in this case is a large corpus of explanations previously scored by domain experts. This corpus (i.e. training set) helps the software "learn" what to look for in the written explanations, and lies at the heart of EvoGrader portal. In order to understand how EvoGrader works, it is necessary to understand some basics about machine learning. The analysis of restaurant ratings will be used as an example to illustrate how machine learning can be applied in order to classify text

Perhaps a restaurant owner is interested in customers' opinions about the meals that are served, and collects a large number of written reviews online. In order to determine how many customers liked and did not like the meals, the restaurant owner must categorize the reviews into groups: e.g., positive reviews and negative reviews. In order to score the reviews a rubric with classification rules would need to be developed. For example, if a review included positive terms such as 'good', 'nice', or 'delicious', then it could be classified into the 'positive review' category. On the other hand, if the review included negative terms such as 'bad', 'unpalatable', or 'disgusting', the review could be classified into the 'negative review' category. Combinations of terms–such as 'not' + key words (e.g., not good, not bad)–could also classified into categories (e.g., 'not good' classified as a "negative review"). With these simple classification rules, the restaurant owner categorizes 300 of the 600 reviews.

Given the amount of time it took to score these 300 reviews, the owner decides to hire a new employee to classify the remaining reviews. In order for the new employee to score the additional reviews, the owner needs to teach the employee how to classify them. There are two options for doing so. First, the owner could show the classification rules (e.g., rubrics) and the lists of terms to the new employee so that he could understand and apply the rules. The problem with this method is that the large number of classification rules makes it difficult to efficiently process the scores. An alternative approach would be to give the new employee all of the previously classified reviews and ask the new employee to infer what classification rules produced the scores (i.e., negatives and positives). Using the second method, the owner only has to provide the scored reviews, and not specify all of the rules and terms used to characterize each review.

Now let us assume that the new employee is replaced by a computer (i.e., machine) that is capable of either using the predefined classification rules, or coming up with new rules based on the pre-classified data. This frames the restaurant review scoring as a machine-based text classification problem. The first method outlined above (i.e., specifying classification rules and terms characteristic of positive and negative restaurant reviews) is known as *rule based text analysis* and has been utilized in prior studies in science education (see [Nehm and Haertig 2012], and [Haudek et al. 2012]). The second method outlined above (i.e., using existing sets of classified data to infer the scoring rules) is how EvoGrader works, and is known as *supervised machine learning* (see [Nehm et al. 2012b], [Ha et al. 2011]). The primary goal of *supervised machine learning* is to discover classification rules given a corpus of classified data and then to apply those rules to score new unlabeled data. Machine learning methods are often used to minimize human effort and interactions while enhancing automation.

Returning to the restaurant example, if the owner decides to take the second approach (i.e., having the new employee use the classified data to infer classification rules and diagnostic text terms) the new employee will need to identify 'positive reviews' and 'negative reviews' and discover key terms diagnostic of each category. In supervised machine learning, the identification of these diagnostic classification terms is known as *feature extraction*. Extracting discriminating and independent features is key to any machine learning process. Next, the employee has to come up with a procedure for building the classification rules based upon the key terms in each review and with respect to each review's category. This procedure is known as *model construction*. The new employee could score a subset of the owner's reviews and compare his classification with the owner's classification to estimate how well his rules worked. In machine learning, this step is known as *model validation*. Once the new employee's classification rules produce identical review ratings as the owner's ratings, he is qualified to classify new sets of reviews. Classification of new reviews is known as *prediction*. In our restaurant example, we can consider the owner to be the *human rater* for the 300 reviews (i.e., the training dataset) and the employee as the "machine" that performs *feature extraction*, *model construction*, *model validation*, and *classification* of new reviews.

Similar to the employee's job of inferring the classification rules that the restaurant owner used to score the reviews as "positive" or "negative," EvoGrader attempts to develop classification rules that predict the presence or absence of particular evolutionary ideas or concepts in a given written explanation. Unlike the restaurant review, where only one classification target is present (positive or negative review), many different classification targets (ideas or concepts) could be present in students' responses to questions about natural selection. Consequently, different scoring models (i.e., classifiers) are needed for each target. Thus, for each concept in students' responses, *feature extraction*, *model construction*, and *model validation* must be performed.

### EvoGrader architecture: back-end training and front-end scoring

Because EvoGrader utilizes supervised machine learning, which in turn relies on discovering classification rules in a corpus of human-classified data, very large pre-scored data sets (text corpora) are required for both system *training* (i.e., learning from existing data) and system *testing* (i.e., examining the strength of classification algorithms). EvoGrader is divided into two major components: (1) a back-end *training* component, and (2) a front-end scoring component (Figure 1). We discuss details of the back-end training and front-end scoring components below.

*Component 1: Backend training.* The construction of scoring models, which takes place in the back-end training part of EvoGrader, is an offline, one-time process. The purpose of the back-end training component is to generate scoring models for *each* concept that will be fed into the front-end component of EvoGrader. Constructing scoring models is a very resource-intensive process; a large corpus is needed for system training, which itself requires large amounts of time, processing power, and memory consumption. Once the scoring models are generated, however, they can be stored permanently and retrieved on-demand by the front-end component. The backend training component includes three parts: (a) *the corpus of human-scored responses*; (b) *the LightSIDE training box*; and (c) *the automatic computer scoring* models (Figure 1).

A very large and diverse human-scored corpus is needed to build classifiers that are capable of accurately predicting the presence or absence of concepts in new samples of written responses. The current version of EvoGrader makes use of 10,270 scored evolutionary explanations written by 2,978 participants (e.g., non-majors, first-year biology majors, senior biology majors, anthropology majors, and experts in evolutionary science [PhD students, post-doctoral fellows, and full-time faculty]). For each of the six key concepts and three naive ideas that have been defined in the evolution education literature (for details, see Nehm et al. 2010), a response was scored by human raters as present (1) or absent (0). Table 1 provides examples of student explanations from the corpus along with human-generated scoring patterns. Two raters (a PhD student in biology education and an evolution expert) scored these explanations and demonstrated acceptable inter-rater human agreement (>0.81 kappa) for all normative scientific ideas and non-normative naive ideas. On average, it took four minutes to score the presence or absence of nine ideas in one explanation. In total, it took approximately 685 hours for the two raters to generate the corpus that EvoGrader utilizes (note that these scored explanations were a byproduct of other research studies, and were not only generated for machine-learning purposes). The final scored corpus used for system training was based on consensus scores between the two human raters.

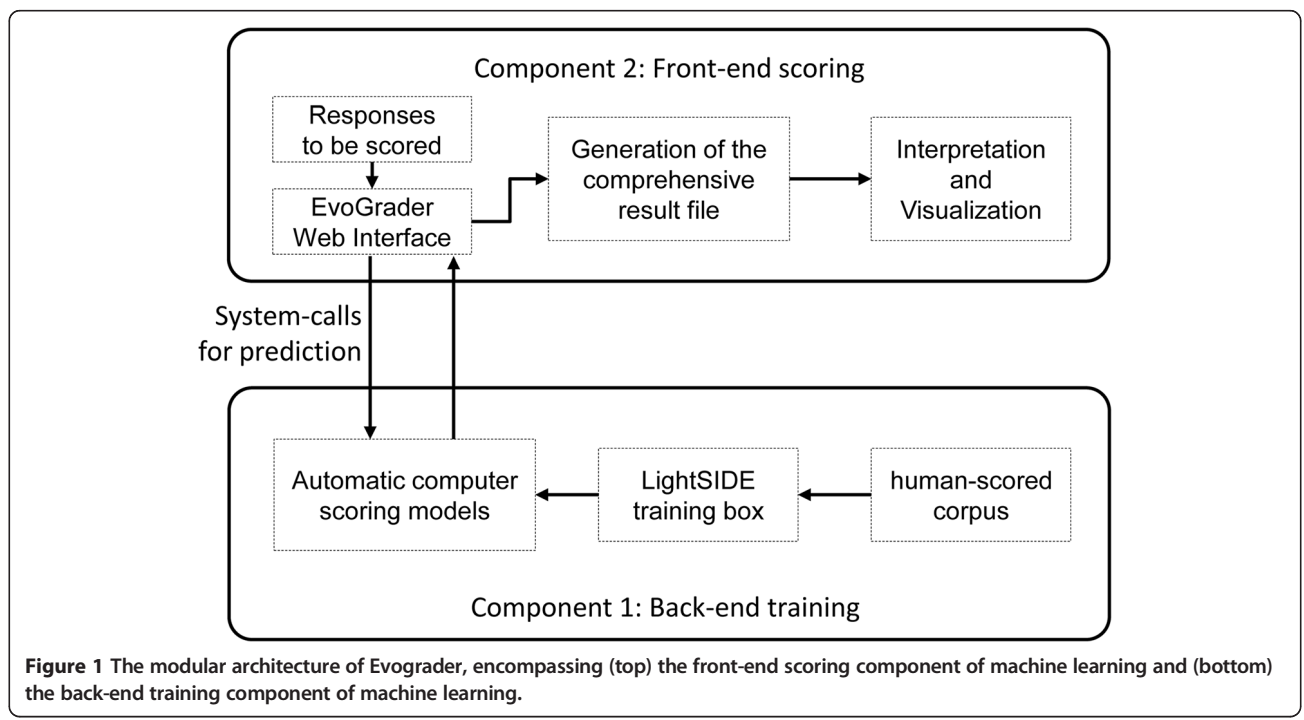

**Figure 1** The modular architecture of Evograder, encompassing (top) the front-end scoring component of machine learning and (bottom) the back-end training component of machine learning.

The back-end of EvoGrader relies on the supervised machine learning tools of LightSIDE Labs' open-source program known as LightSIDE (Mayfield *et al.* 2013). LightSIDE performs three important functions in EvoGrader: *feature extraction*, *model construction*, and *model validation*. Using the human-scored corpus as a training set, algorithms are sought that yield nine unique scoring models (corresponding to the nine concepts). Offline, using the human-scored corpus as a training set, LightSIDE develops scoring models for each of the nine concepts (see Table 1). The goal of this process is to find classifiers that fit the training dataset maximally, and subsequently *predict* scores on new datasets with high accuracy. Two approaches are used to determine whether these goals are met: cross validation and human expert review.

To build the nine scoring models offline, the human-scored corpus is imported into the LightSIDE application. *Feature extraction* in LightSIDE begins by turning each response in the corpus into a "bag of words" (Harris 1954). This approach simplifies text representation and reduces the dimensionality of the feature space. An inherent assumption in the bag of words representation is that words are considered independent of their place in the text. All of the words that have appeared at least once in the corpus are entered into a 'corpus dictionary.' Highly frequent neutral words (e.g., "the, of, to"), punctuations (e.g., ","), and very infrequent words (e.g., "shall") are removed from the corpus in order to minimize noise. The remaining words are significant and as long as they appear at a frequency above a cutoff value they will be included in the dictionary. Stemming may also be used, which treats words with the same stem (e.g., the stem "adapt" is common to "adapt", "adapting", and "adaptation") as a single word. Bigram modeling (Cavnar and Trenkle 1994) may also be used and allows for the creation of double-word terms (e.g., "passing on", "differential survival") in the dictionary.

It is important to emphasize that different concepts typically require different *feature extraction* settings (e.g., unigrams vs. bigrams, stemming vs. non-stemming, low vs. high cutoff values) (see Ha *et al.* 2013). For example, two bigram features – 'had_to', 'so_that'— are comprised of highly frequent neutral words (e.g., had, to, so, that); nevertheless, for some concepts–such as the non-normative concept of evolutionary needs/goals–they are crucial for building effective scoring models. The statistically optimized conditions for *feature extraction* for all nine scoring algorithms in EvoGrader are shown in Table 2.

After building the corpus dictionary, each written explanation is converted into a vector of zero/nonzero values. These vectors are called *feature vectors* (Asadi and Lin). Each of the student responses in the corpus is modeled as a sparse vector. The non-zero values for each vector are the frequencies of the words present in the corresponding response, and the zero values represent all of the other words in the corpus dictionary that did not appear in that particular response. The outputs of the feature extraction stage are these feature vectors.

**Table 1 Selected examples of students' written explanations of evolutionary change and corresponding human scores from the training set of EvoGrader**

| Student's responses to ACORNS or ACORNS-like items | Normative Scientific idea | | | | | | Non-normative Naive idea | | | Reasoning model type |
|---|---|---|---|---|---|---|---|---|---|---|
| | Variation [V] | Heredity [H] | Competition [C] | Limited resources [R] | Different survival [D] | Non-adaptive idea [NA] | *Need/ goal [N]* | *Use/ disuse [U]* | *Adapt/ acclimation [A]* | |
| *An elm tree may have had some seeds that were shaped a little differently from the others* [V], which allowed them to land on the ground farther away from the parent tree. *Those seeds may have germinated and sprouted better than the traditionally shaped seeds* [D] because they didn't land right under the parent tree and then had to *compete* [C] for *nutrients and sunlight* [R]. If the shape of winged seeds was genetic, *those seeds could pass those genetic changes on to the seeds they produce* [H]. *Over long periods of time, natural selection selected for those seeds that were more wing shaped* [D], until all of the seeds that successfully grew to adults all were wing shaped. | 1 | 1 | 1 | 1 | 1 | 0 | 0 | 0 | 0 | Pure scientific model |
| The birds may have colonized a new area where *predators are absent* [R]. So the wings no longer provided an advantage in terms of escaping predation. At the same time the birds may be exploiting new sources of food in the water. Any *mutation* [V] that allowed the wing to successfully function more like a flipper would be advantageous (even if it had a deleterious effect on flight). *These birds would have higher reproductive success* [D]. *If the mutation is heritable* [H] it will increase in frequency in the population until it becomes fixed. | 1 | 1 | 0 | 1 | 1 | 0 | 0 | 0 | 0 | Pure scientific model |
| *The cacti without spines had to have developed spines over time* [A] due to changes in its environment such as *to prevent itself* [N] *from harm from other threats* [R]. *This trait of having spines therefore became favorable* [D] and *the trait will be passed down by generation to the next offspring and so on* [H]. Therefore, the cacti today all have spines. | 0 | 1 | 0 | 1 | 1 | 0 | 1 | 0 | 1 | Mixed model |

Table 2 Concept types, names, and descriptions of EvoGrader scoring models

| Concept type | Concept name | Concept description | Methods used to optimize scoring models |
| --- | --- | --- | --- |
| Normative scientific idea | Variation | The presence and causes (mutation/recombination/sex) of variation | Unigram, Stemming, Removing Stopwords, Removing misclassified data |
| | Heritability | The heritability of variation (The degree to which a trait is transmitted from parents to offspring) | Unigram, Bigram, Stemming, Removing Stopwords, Removing misclassified data |
| | Competition | A situation in which two or more individuals struggle to get resources that are not available to everyone | Unigram, Stemming, Removing Stopwords |
| | Limited resources | Limited resources related to survival/reproduction, such as food and predators, and reproduction (such as pollinators) | Unigram, Stemming, Removing Stopwords, Removing misclassified data |
| | Differential survival/ | The differential reproduction and/or survival of individuals | Unigram, Bigram, Stemming, Removing Stopwords |
| | Non-adaptive idea | Genetic drift and related non-adaptive factors contributing to evolutionary change | Unigram, Stemming, Removing Stopwords |
| Non-normative naïve idea | Adapt/acclimation | Adjustment or acclimation to circumstances (which may subsequently be inherited) | Unigram, Bigram, Stemming, Removing Stopwords, Removing misclassified data |
| | Need/goal | Goal-directed change; needs as a direct cause of evolutionary change | Unigram, Bigram, Stemming, Removing misclassified data |
| | Use/disuse | The use (or lack of use) of traits directly causes their evolutionary increase or decrease | Unigram, Bigram, Stemming, Removing Stopwords |

Once *feature extraction* has been completed, *model building* and *validation* begin. In order to build scoring models, decision functions must be generated that are based on running the learning algorithm on the feature vectors. The objective of this process is to build binary classifiers for each concept that are capable of accurately predicting the presence or absence (1/0) of a concept in a given explanation. Unpublished research (Nehm, unpublished data) suggests that Sequential Minimal Optimization (SMO) (Platt 1999) is the most effective algorithm for the corpus used in EvoGrader.

Each binary classifier represents a decision function for one of the concepts. Therefore, SMO training is performed nine times (once for each concept). For each concept, inputs to the SMO training algorithm include feature vectors and corresponding human scores for that concept. SMO training is an iterative process that repeatedly chooses weight factors for the dictionary words based on the feature vectors and human scores for that concept. These iterations (again, for each individual concept) will continue until the binary classifier that is generated at the end of the iteration is able to classify all of the responses correctly with certain error threshold. Training the classifiers is a one-time process.

To validate each of these nine classifiers (for each of the nine concepts), ten-fold *cross-validation* is performed on each concept model. The cross-validation process involves segregating the corpus into $k$ subsets (e.g., ten-fold refers to 10 subsets) and performing model construction with $k - 1$ subsets (e.g., nine subsets if ten-fold cross-validation is used); model validation is performed with the last subset. The ten-fold cross-validation approach used in EvoGrader repeats this process 10 times and averages the results so that each subset can be validated once.

The training corpus may contain the data misclassified by the cross-validation process (Muhlenbach et al. 2004). Although the reasons for the misclassified data are usually the mislabeling of the data or spelling errors, the reasons are not apparent in many cases because of the complexity of the machine learning algorithm. In such cases, we can remove the misclassified data from the original training corpus and re-train the scoring algorithm using the new corpus (Muhlenbach et al. 2004).

After *feature extraction* and *model construction* are completed for the nine binary classifiers, their performance is compared to performance thresholds (90% accuracy, and 0.8 kappa coefficients). If they fail, they are subjected to feature refinement. For example, using bigrams *and* unigrams might produce better performance than using unigrams alone (see Ha et al. 2013, for details on this process). Once the scoring models meet these benchmarks, they are saved and used by the front-end scoring component of EvoGrader.

*Component 2: Front-end scoring.* The scoring models generated in the back-end component of EvoGrader are used in the front-end component to classify each new response as "present" or "absent" for each of the nine concepts. To perform prediction, each new response is converted into a vector representation (as discussed above). Using the weights generated previously in the SMO training step, the system calculates whether the vector belongs to the "absent" class or to the "present" class.

The scoring component of EvoGrader is supported by Amazon's Elastic Cloud service (EC2). EC2 provides resizeable, on-demand computation capacity in the 'cloud'. In terms of scalability of the hosting service, EC2 is an appropriate fit for EvoGrader because 'bursty' http traffic and sudden processing loads for scoring large datasets is anticipated to be common; it is difficult to predict when users will upload data files for analysis. Hosting EvoGrader on a virtual private cloud provides a flexible web service environment that is capable of rescaling processing and memory power to maintain a reasonable response time for all of the users irrespective of the number of concurrent users.

The front-end scoring component of EvoGrader includes: (a) *system-calls for prediction and generation of the comprehensive result file,* and (b) *interpretation and visualization.* We discuss each of these components in turn.

*System-calls for prediction and generation of the comprehensive result file.* EvoGrader requires users to format student response data prior to upload into the web portal. Step-by-step video instructions are provided in the web portal (and written instructions are included in the Additional file 1). In brief, student identifiers and typed explanations are pasted into separate cells in a spreadsheet and the file is then saved in .csv (comma separated values) format. Response files can contain an unlimited number of typed explanations to up to eight different ACORNS items in one run (unlimited numbers of runs can be performed). File setup usually takes less than five minutes.

After upload to the EvoGrader portal, the system performs a round of preprocessing and verifies that the response file has been formatted properly. Then, for each item, scoring model files from the back-end training component are used to execute online scoring using LightSIDE. As noted above, LightSIDE is a standalone application, and therefore its prediction engine cannot service clients directly over the web. EvoGrader's front-end performs this function. The application server generates a system call to invoke the local LightSIDE predictor. This system call will send an interrupt message to the operating system on the application server, and will initiate the scoring process using the parsed response file and the pre-constructed scoring models. In files that include responses to multiple items, EvoGrader generates temporary files for each item and passes them sequentially to the LightSIDE prediction engine. The automated scoring process usually takes less than an hour and the results are stored on the application server. As soon as the final temporary file corresponding to the last item is scored, the application server merges all of the results to produce a unified comprehensive result file for all of the items included in that run. At this stage, another system call from LightSIDE invokes loading of a java servlet that provides a downloadable result file.

*Interpretation and visualization.* The downloadable result file contains raw scoring results for all of the concepts and reasoning models. However, instructors do not need to download any files to discover actionable insights about their student response file. EvoGrader automatically generates charts and tables to help instructors interpret and visualize response patterns (Figure 2). Using Java servlet technology on the application server, as soon as the result file is generated, in the background a parallel thread starts executing a general analysis of the results to generate required data for tables and charts. This parallel processing enhances resource and time utilization. Bar charts, pie charts, and bubble charts, visually represent types of reasoning models and concept use patterns in the sample (See Figure 2).

### Using the EvoGrader system

Although it is apparent that the computational machinery that EvoGrader employs is complex (see above), using the system is easy. Users only need to know how to: (1) use spreadsheets, (2) upload a CSV file to the web, and (3) use an online web application. Complete step-by-step instructions are included in video format on the EvoGrader website, and written instructions are included in the Additional file 1 for this paper. Overall, the system was designed to be usable to anyone with very basic computer skills.

### Research questions

Prior work has demonstrated the utility of machine-learning and text analysis methods for scoring written evolutionary explanations (e.g., Beggrow et al.; Ha and Nehm 2012; Nehm and Haertig 2012). The present study goes beyond this prior work to analyze the efficacy of new scoring models derived from a much larger and more diverse human-scored corpus and optimized for use in an online environment (i.e., in the EvoGrader online portal). In addition to testing how well EvoGrader scores ACORNS instrument items, we also examine how well EvoGrader scores "ACORNS-like" items (i.e., not the 86 items designed for use in EvoGrader, but similar open-response evolution items). Because different assessment prompts and contexts have been shown to influence students' ideas (and the corresponding language used to express these ideas, see Nehm and Ha 2011), it is an open question as to whether EvoGrader might be able to effectively grade written responses to ACORNS-like questions developed by other researchers. Some instructors may want to use their own short-answer questions, or they may have already collected students' essays relating to natural selection before they became aware of the ACORNS instrument. For these reasons, it would be useful to know how well the system works with such responses. In short, the empirical questions addressed in this study are: (1) How well do the new scoring models in the EvoGrader portal perform compared to trained human scorers? And (2) How well does the EvoGrader portal perform on ACORNS-like items? In order to compare EvoGrader performance to human performance in these two contexts, several different methods are employed using two new student response corpora.

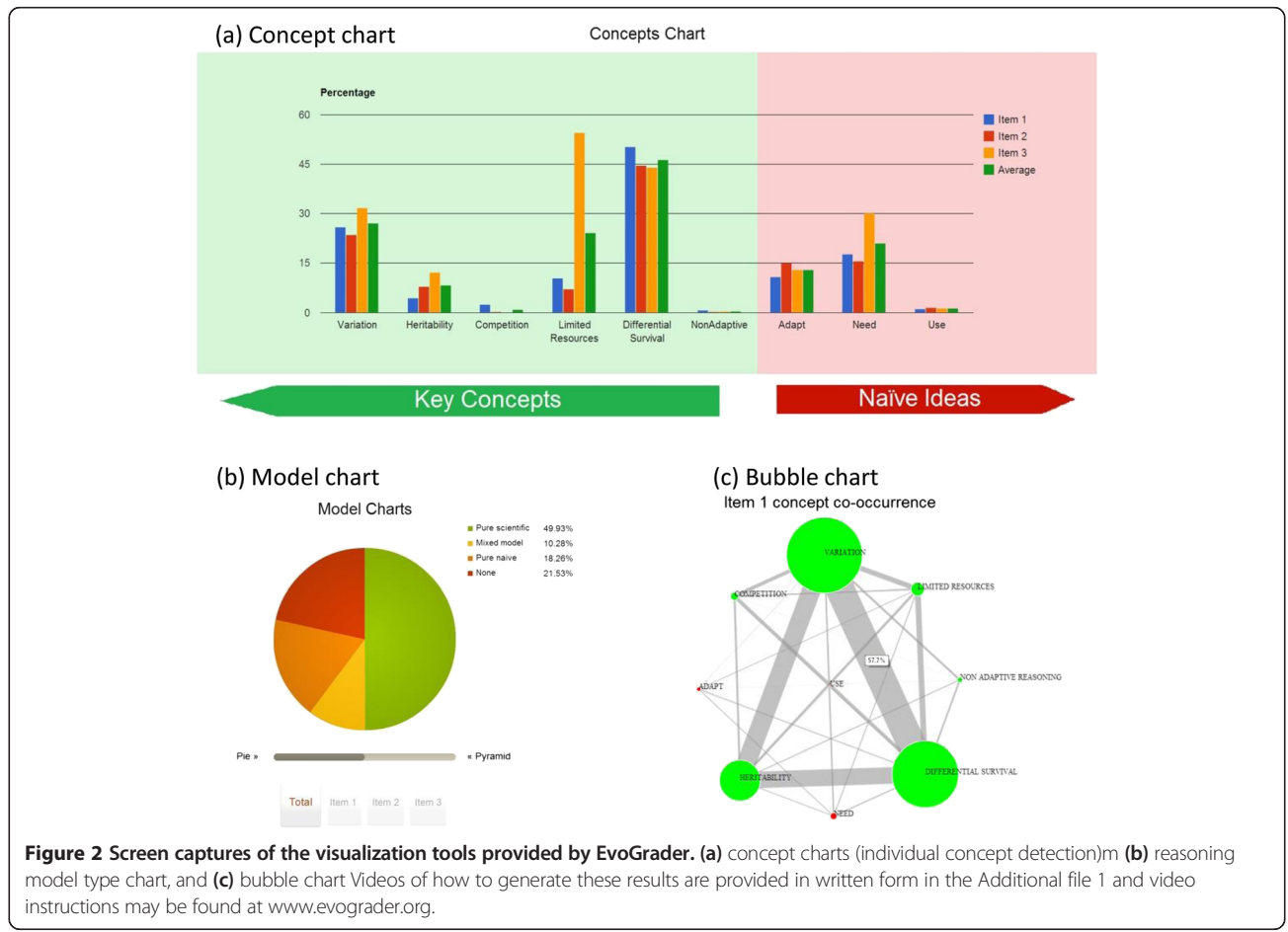

**Figure 2 Screen captures of the visualization tools provided by EvoGrader. (a)** concept charts (individual concept detection)m **(b)** reasoning model type chart, and **(c)** bubble chart Videos of how to generate these results are provided in written form in the Additional file 1 and video instructions may be found at www.evograder.org.

## Methods

### Human vs. computer scoring of evolutionary explanations: concepts and models

In order to evaluate EvoGrader's scoring performance, EvoGrader-generated scores were compared to the human-generated scores. Two different categories of scores were compared: first, those for six normative scientific concepts (*variation, heritability, competition, limited resources, differential survival/reproduction,* and *non-adaptive ideas* (e.g., genetic drift) and those for three non-normative ideas (*needs and goals* [teleology], *use and disuse* ['use-inheritance], and *adapt/acclimation*) (see Table 2 for detailed descriptions of these concepts). All of these ideas have been commonly found in research on student thinking about evolution (e.g., Bishop and Anderson 1990; Nehm and Reilly 2007).

In addition to studying concept use patterns in the written responses, we also examined students' overall reasoning patterns (i.e., holistic models). Specifically, we used the cognitive models outlined by Nehm et al. (2009) to categorize each student response into one of four categories: a scientific model (including only normative scientific ideas), a mixed model (including both scientific and naïve ideas), a naïve model (including only non-normative naive ideas), or no model (rephrasing the question, providing extraneous information, or not answering the question directly).

The scoring rubrics of Nehm *et al.* (2010) were used to guide the production of human-generated scores. Detailed examples of student responses, and how to score them, are included in this 40-page rubric set. Two trained human raters who demonstrated sufficient inter-rater agreement (>0.80 kappa for all normative scientific ideas and non-normative naive ideas) produced these scores. Consensus scores between the two raters were used for all statistical comparisons to EvoGrader.

### Statistical tests of EvoGrader's scoring performance

The scoring efficacy and accuracy of EvoGrader were tested using correspondence magnitudes (i.e., inter-rater agreement) between human and computer scores. Several statistical methods were used: Cohen's kappa, raw percentage agreement, precision, recall, $F_1$ scores, and Spearman's rank correlation. Cohen's kappa accounts for chance agreements between raters, and is a widely-used measure quantifying inter-rater reliability in education and medical research. In our study, the two raters were EvoGrader and a trained human scorer. Although different studies have suggested different cutoff scores for Cohen's kappa, the benchmarks suggested by Landis and Koch (1977) and Fleiss (1971) were used in the evaluation of EvoGrader performance: 'almost perfect': 0.81-1.00 (Landis and Koch 1977) and 'excellent': 0.75 – 1.00 (Fleiss 1971) (see Bejar (1991) and Ha et al.([Ha et al. 2011]) for additional details about these tests).

Raw agreement percentages are also reported (the percentage of total agreements between EvoGrader and the human rater). Note that this raw measure does not account for chance agreements, but it provides a very intuitive metric of correspondence.

Precision and recall are widely used measures in information retrieval studies (Ha *et al.* 2013; Su 1994). *Precision* indicates the percentage of correct predictions among total positive predictions, whereas *recall* indicates the percentage of correct predictions among total positive cases. Precision refers to how frequently EvoGrader overestimates correct responses. For example, precision will decrease when the scoring model predicts that a particular concept is shown in the student's response when in fact it is not. In contrast, recall refers to how many responses containing correct concepts were appropriately classified. Therefore, recall will decrease when the scoring models fail to detect a concept. Precision and recall exhibit a trade-off relationship; when one increases, the other decreases. Consequently, a composite measure of precision and recall known as $F_1$ is often used (i.e., the harmonic mean of precision and recall).

Spearman's rank correlation coefficients were also used to calculate correspondence between two different 'holistic' measures: total key concept scores and total naïve idea scores. Although Pearson correlation coefficients have been used to quantify such correspondence in previous studies (Beggrow et al.; Ha and Nehm 2012), Spearman's rank correlation is more appropriate in the present study because the range of scores for a single item was small and the distribution was not normal. Quality benchmarks for Pearson correlation coefficients have been established in bioinformatics; many bioinformatics studies consider Pearson correlation coefficients > 0.9 to be 'nearly identical' (e.g., Sato *et al.* 2005; Zhu *et al.* 2002) and so we adopted this benchmark for our Spearman's rank correlation tests.

### Participant samples used to test scoring performance

Two new student response corpora were used to compare EvoGrader performance to human-rater performance. The first corpus included 1100 student answers written in response to 86 different ACORNS items. This sampling strategy was used to produce a conservative estimate of EvoGrader performance. All of these responses were from students enrolled in introductory biology classes at one public Midwestern university.

The second corpus was produced in response to an "ACORNS-like" item that was developed by a researcher as part of an unrelated study. The second corpus (n = 1100) contained students' responses to a question about the evolution of orchid leaves. These data were collected from undergraduates at 59 different institutions. In sum, 2200 written explanations were independently scored for the presence or absence of evolutionary concepts by both EvoGrader and by humans.

### Results

Table 3 illustrates five correspondence measures (i.e., kappa values, agreement percent, precision, recall, and $F_1$ scores) for the six key concepts (i.e., *variation, heritability, differential survival, competition, limited resources,* and *non-adaptive ideas*). As shown in Table 3, the kappa values for all six key concepts exceeded the 'almost perfect' (>0.81) level, and raw percentage agreement levels reached or exceeded 95% for ACORNS responses. In contrast, for ACORNS-like responses, kappa values for the *limited resources* concept did not reach the 'almost perfect' (>0.81) level (0.806). However, kappa values for the other five key concepts exceeded the 'almost perfect' (>0.81) level (See Figure 3).

For naïve ideas, the kappa values and agreement percentages for *needs/goals* and *use/disuse* exceeded the 'almost perfect' (>0.81) kappa value and nearly reached or exceeded the 95% agreement level. However, the naïve ideas of 'adapt/acclimation' did not reach the 0.75 kappa value that Fleiss (1971) considered to be an excellent kappa value. Nevertheless, the raw agreement percentage for *adapt/acclimation* almost reached 95%. For the naïve ideas detected in ACORNS-like responses, the kappa values and agreement percentages for *needs/goals* also exceeded the 'almost perfect' (>0.81) kappa value and nearly reached the 95% agreement level. However, the naïve ideas of use/disuse and adapt/acclimation did not reach the 0.70 kappa level.

Prior to reporting the human-EvoGrader kappa and raw agreement correspondence measures for the four evolutionary reasoning models (e.g., scientific models, mixed models, naïve models, and no models), it is important to note that whereas binary coding characterized the concept presence/absence data reported above, quaternary (4*4) coding characterized the reasoning model data. Consequently, the quaternary human-computer model agreement will be less probable than the binary concept agreement. Nevertheless, kappa agreement values for model types were strong and exceeded 0.81; agreement percentages were also robust at 88.4% (See Figure 4). On the other hand, the kappa and agreement percentages for ACORNS-like items were lower than those for ACORNS responses (kappa: 0.766, agreement percentage: 84.8%).

**Table 3 Performance measures for EvoGrader for ACORNS (n = 1100 written explanations) and ACORNS-like written responses (n = 1100 written resppnses) (n = 2200 total written explanations)**

| Concept type | Concept | Concept frequency (%) | | Kappa | | Agreement (%) | | Precision | | Recall | | $F_1$ | |
|---|---|---|---|---|---|---|---|---|---|---|---|---|---|
| | | ACORNS | ACORNS-like | ACORNS | ACORNS-like | ACORNS | ACORNS-like | ACORNS | ACORNS-like | ACORNS | ACORNS-like | ACORNS | ACORNS-like |
| Normative scientific idea | Variation | 35.3 | 21.5 | 0.903 | 0.822 | 95.6 | 94.4 | 97.0 | 93.9 | 90.5 | 78.8 | 93.6 | 85.7 |
| | Heritability | 14.7 | 22.3 | 0.879 | 0.852 | 97.1 | 95.2 | 95.1 | 97.5 | 84.6 | 80.4 | 89.5 | 88.1 |
| | Competition | 1.6 | 2.2 | 0.971 | 0.932 | 99.9 | 99.7 | 100.0 | 100.0 | 94.4 | 87.5 | 97.1 | 93.3 |
| | Limited resources | 20.5 | 22.4 | 0.944 | 0.806 | 98.2 | 93.9 | 96.4 | 99.4 | 94.7 | 73.2 | 95.5 | 84.3 |
| | Differential survival | 46.2 | 55.8 | 0.855 | 0.851 | 92.8 | 92.6 | 93.2 | 94.6 | 91.1 | 92.0 | 92.1 | 93.3 |
| | Non-adaptive idea | 3.0 | 0.1 | 0.984 | 1.000 | 99.9 | 100.0 | 100.0 | 100.0 | 97.0 | 100.0 | 98.5 | 100.0 |
| Non-normative naïve idea | Need/Goal | 22.8 | 29.0 | 0.849 | 0.871 | 94.7 | 94.8 | 89.4 | 95.2 | 87.3 | 86.5 | 88.3 | 90.6 |
| | Use/Disuse | 4.1 | 4.3 | 0.848 | 0.653 | 98.8 | 97.4 | 86.4 | 72.5 | 84.4 | 61.7 | 85.4 | 66.7 |
| | Adapt/Acclimation | 10.6 | 9.7 | 0.718 | 0.651 | 94.7 | 95.1 | 76.1 | 94.9 | 73.5 | 52.3 | 74.8 | 67.5 |

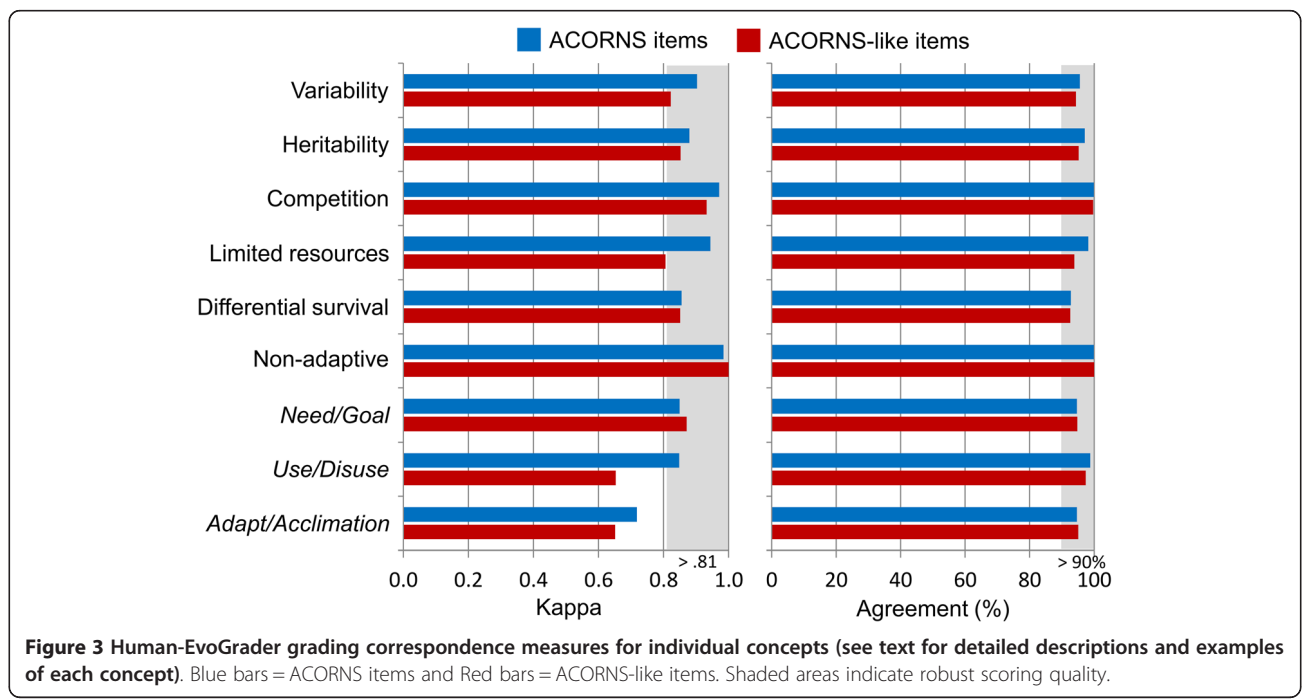

**Figure 3 Human-EvoGrader grading correspondence measures for individual concepts (see text for detailed descriptions and examples of each concept).** Blue bars = ACORNS items and Red bars = ACORNS-like items. Shaded areas indicate robust scoring quality.

Spearman's rank correlations of total key concept and naïve idea scores for the ACORNS responses were 0.927 ($p << .01$) and 0.835 ($p << .01$), respectively. On the other hand, Spearman's rank correlation total scores for key concepts and naïve ideas for the ACORNS-like responses were slightly lower at 0.890 ($p << .01$) and 0.837 ($p << .01$), respectively.

In sum, EvoGrader performed very well at the concept and reasoning model level for ACORNS items, and did fairly well when provided with responses to "ACORNS-like" items.

## Discussion
### The scoring efficacy of EvoGrader
In undergraduate biology education, few formative assessment systems exist for helping students practice building scientific explanations using core ideas (such as evolution and natural selection) (see National Research Council 2012). The EvoGrader system is one of the first non-commercial online tools to use machine learning to evaluate evolutionary explanations. Our research findings demonstrate the efficacy of the system using responses to both ACORNS and ACORNS-like assessment items. EvoGrader successfully scores and reports the key concepts, naïve ideas, and evolutionary reasoning models contained within students' written explanations of evolutionary changes within an hour (or slightly longer for > 5000 written explanations).

Compared to commonly used benchmarks of statistical correspondence, EvoGrader-generated key concept scores are "almost perfect" (kappa > 0.81) and "nearly identical" (correlation > 0.9) to human-generated scores. Specifically, the kappa coefficients for all six key concepts (e.g., *variation, heritability, competition, limited resource, differential survival/reproduction, non-adaptive idea*) exceeded 0.81, and those for *competition*, *limited resources*, and *non-adaptive ideas* exceeded 0.91. The correlation between EvoGrader-generated and human-scored total key concept scores was 0.927.

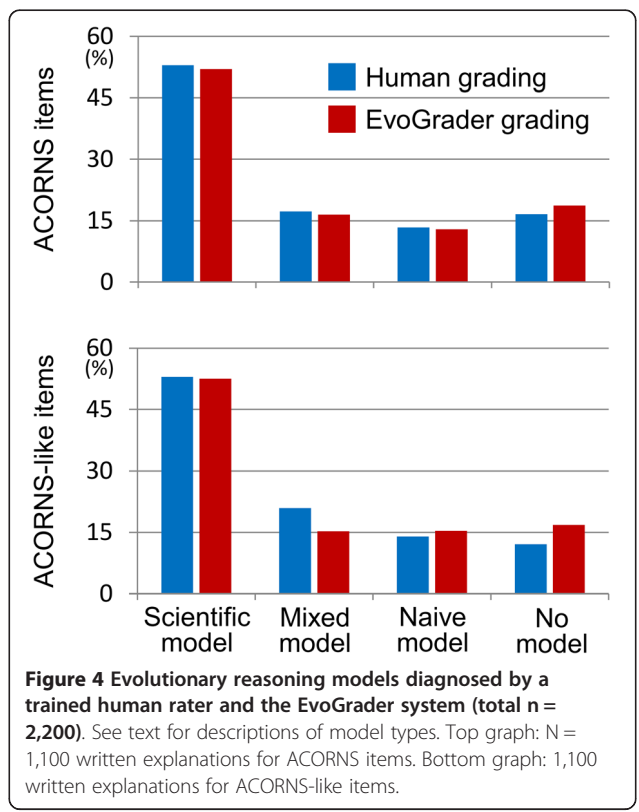

**Figure 4 Evolutionary reasoning models diagnosed by a trained human rater and the EvoGrader system (total n = 2,200).** See text for descriptions of model types. Top graph: N = 1,100 written explanations for ACORNS items. Bottom graph: 1,100 written explanations for ACORNS-like items.

EvoGrader scoring models for naïve ideas displayed robust but slightly lower agreement compared to the results for key concepts. EvoGrader detected the naïve ideas of *needs/goals* (0.849) and *use/disuse* (0.848) above the benchmark of kappa (0.81). In contrast, *adapt/acclimation* had a kappa score of 0.718 and a raw agreement of 94.7%. Despite EvoGrader's slightly lower performance on this naïve idea, the correlation between human- and EvoGrader-generated total naïve idea scores was 0.835.

It should be noted that the frequency of each concept differs in the response corpus (see Table 3). Compared to the most common naïve idea (*needs/goals*, 25.9% of total responses), the percentage of *adapt/acclimation* ideas was only 10.2% of total responses. Thus, the weaker kappa performance of the scoring model for *adapt/acclimation* is due, in part, to the fact that it was not very common in our particular response corpus.

Finally, perhaps the most important measure of students' evolutionary explanations is a holistic judgment of their quality. Kappa coefficients for the four reasoning models all exceeded the "almost perfect" level (kappa = 0.821). Given these results, EvoGrader-generated scores can be considered comparable to human holistic judgments of the quality of evolutionary explanations.

EvoGrader can be used to score students' written responses to ACORNS-like instruments, although scoring quality is not as robust. This result is in line with what we know about machine-learning methods (see Introduction). Given that (a) students' responses are known to be strongly influenced by the prompts provided to them and the item features in these prompts (see Nehm and Ha 2011) and (b) students' language is used to build machine-learning scoring models, the success of EvoGrader's scoring algorithms will be dependent on the training corpus. Our empirical results support this line of reasoning and demonstrate that EvoGrader showed better performance for ACORNS responses relative to ACORNS-like responses. Nevertheless, with larger training corpora from more diverse items, it is possible that EvoGrader could be improved to be less dependent on item features. Indeed, our results show that EvoGrader accurately detected many key concepts and naïve ideas in ACORNS-like items. This is a promising finding that suggests that more broad-based scoring models are possible.

### EvoGrader in the classroom

EvoGrader's robust and rapid performance suggests that it could be a useful assessment tool in advanced high school and undergraduate biology classes. Three main uses for the EvoGrader system include: (1) a diagnostic pre-assessment tool to help classroom teachers and university faculty plan evolution instruction; (2) a homework/instructional activity for use during an evolution unit or course; and (3) as a value-added measurement tool for quantifying learning gains in an instructional unit or course. At present, the EvoGrader system is not meant for use as a high-stakes test; that is, the system was developed as a tool to help instructors better understand student thinking and reasoning about evolution and how that understanding changes over time.

Diagnostic pre-tests can be of great value in planning and designing instruction: they help to identify the knowledge, practices, and naive ideas that students utilize to explain evolution, and reveal the quality of explanation and communication skills. Course management systems or free survey software may be used to gather student responses to ACORNS items before a class begins. Depending upon the student response patterns, differentiated instructional activities can be designed to address common naive ideas, explain key concepts, or illustrate causally robust scientific explanations. Diagnostic pre-tests help to target instruction on particularly problematic topics characteristic of a population of students.

EvoGrader may also be used as an instructional tool to help students practice explaining evolutionary change across a variety of scenarios (e.g., antibiotic resistance in bacteria, plant leaf size change, mammal digit evolution). It may also be used to help students consider what features characterize a robust scientific explanation, and practice developing an evolutionary explanation (see National Research Council 2012).

EvoGrader may also be used as a tool to measure the ways in which students' evolutionary reasoning changes before and after a course or instructional unit. Currently, most concept inventories require the use of the same items before and after an intervention. Naturally, one would expect improvements given that the same items have been administered. With more than 80 items in the EvoGrader system, many alternative items exist for pre-post testing.

Future work planned for the EvoGrader system includes (1) more rapid grading, so that open-ended "clicker questions" can be graded in real time; (2) the improvement of scoring models for existing concepts; and (3) the expansion of concepts that the system can reliably score.

### Adapting the software architecture of EvoGrader to other domains

Although EvoGrader was designed to function using written evolutionary concepts (particularly natural selection), a web application with very similar characteristics (and a different training corpus) could be built to serve educators interested in automated analysis of other concepts in other domains (e.g., phylogenetics, macroevolution, speciation). Indeed, EvoGrader can be considered to be a customizable educational tool. EvoGrader serves users in two distinct ways (see Figure 1): a scoring track and a training track. As mentioned earlier, the scoring track is openly accessible to all instructors to upload their response files. The system subsequently generates scores, performs analyses, and displays visualizations of results. The training track is the customizable side of the system, and it is only accessible for the users with administrative privileges. Such users are allowed to update the training dataset, and are able to reconstruct the scoring models based on their own human-scored training sets. This feature adds flexibility to EvoGrader's core functions.

For various reasons, the training dataset may need to be modified over time. This is useful in cases in which: additional human-scored data have become available, new concepts have been identified and scored, changes have been made in scoring rubrics, improved classification approaches have been developed, or misclassified data need to be removed from the corpus. Very similar to the scoring track scenario, in the training track (after uploading the new training set), a system call is generated on the application server to invoke the LightSIDE model building engine with the proper feature set configurations (see Figure 1). Therefore, LightSIDE will use the SMO training iteratively over the new training set to generate new scoring models. After building these scoring models using the training track, any response file that is provided by instructors will be scored, analyzed, and interpreted using the new scoring models. In sum, although EvoGrader is a specialized tool, its architecture is generic with respect to the training data.

Science educators who are interested in developing an automated scoring tool in their own domain can follow the same approach and mirror EvoGrader's architectural configuration (Figure 1). The first step is developing clear scoring rubrics and then gathering and scoring a sufficiently large and diverse corpus of responses. The next step is to identify the supervised machine learning algorithm that best characterizes domain-specific features. Validation measures (such as those used with EvoGrader, see above) may be used to empirically evaluate the performance of the scoring algorithm (see Beggrow et al. in press; Ha et al. 2011; Nehm et al. 2012b for examples). LightSIDE is the core of EvoGrader, and is an open-source tool that can be modified; it provides different classification algorithms and validation metrics suited to many different purposes and may be used for any type of machine-learning purpose. Once robust scoring models have been developed using the tools in LightSIDE, a web application similar to EvoGrader's front-end component (see Figure 1) can be developed with a focus on concept-specific representations and interpretations. In short, the basic architecture of EvoGrader provides a low-cost template for the development of other automated scoring tools in other domains.

## Conclusions

While it is clear that technological advances and the demand for richer and more informative types of tests have begun to change the field of educational assessment in the United States (Duncan 2013), nearly all evolution assessments remain in multiple-choice formats. Educators must begin to embrace new tools and technologies and use them to develop more meaningful measures of evolutionary thinking and reasoning (National Research Council 2012). EvoGrader is one small step in this important direction.

## Additional file

**Additional file 1: User guide.**

**Abbreviations**
ACORNS: Assessing COntextual reasoning about natural selection; SMO: Sequential minimal optimization.

**Competing interests**
The authors declare that they have no competing interests.

**Authors' contributions**
KM was responsible for computer system design, MH was responsible for analysis of data, and RHN was responsible for study design. MH and RHN were equally responsible for data collection. All authors were equally responsible for the drafting of the manuscript. All authors read and approved the final manuscript.

**Acknowledgements**
We thank National Science Foundation REESE grant 0909999 and NSF TUES grants 1338570 and 1322872 for funding portions of this work. Any opinions, findings, and conclusions or recommendations expressed in this publication are those of the authors and do not necessarily reflect the views of the NSF. Please see www.evograder.org for additional information.

**Author details**
[1]Department of Computer Science and Engineering, The Ohio State University, 2015 Neil Avenue, 43210 Columbus, OH, USA. [2]Center for Science and Mathematics Education, Stony Brook University, 092 Life Sciences Building, 11794 Stony Brook, NY, USA. [3]Center for Science and Mathematics Education, and Department of Ecology and Evolution, Stony Brook University, 092 Life Sciences Building, 11794 Stony Brook, NY, USA.